%% file: main.tex
\documentclass[
    a4paper,
    twoside,
]{article}

\usepackage{epsfig}
\usepackage{subcaption}
\usepackage{calc}
\usepackage{amssymb}
\usepackage{amstext}
\usepackage{amsmath}
\usepackage{amsthm}
\usepackage{multicol}
\usepackage{pslatex}
\usepackage{apalike}
\usepackage{algorithm2e}
\usepackage[bottom]{footmisc}

\usepackage{hyperref}
\usepackage[capitalize]{cleveref}
\usepackage{csquotes}
\usepackage{multirow}
\usepackage{siunitx}
\usepackage{booktabs}
\usepackage{nicefrac}
\usepackage{xcolor}

\definecolor{HSG-Green}{RGB}{0,128,47}
\definecolor{HSG-Darkgreen}{RGB}{10,94,45}
\definecolor{HSG-Beige}{RGB}{225,215,195}
\definecolor{HSG-Blue}{RGB}{115,165,175}
\definecolor{HSG-Red}{RGB}{235,105,105}
\definecolor{HSG-Yellow}{RGB}{255,240,75}
\definecolor{HSG-White}{RGB}{255,255,255}
\definecolor{HSG-Black}{RGB}{0,0,0}

\usepackage{tikz}
\usetikzlibrary{arrows}
\usetikzlibrary{backgrounds}
\usetikzlibrary{calc}
\usetikzlibrary{positioning}

\usepackage{pgfplots}
\pgfplotsset{compat=1.17}
\usepgfplotslibrary{external}
\hypersetup{%
    pdftoolbar=true,        
    pdfmenubar=true,        
    pdffitwindow=false,     
    pdfstartview={FitH},    
    pdftitle={Reducing the Transformer Architecture to a Minimum},    
    pdfauthor={Bernhard Bermeitinger, Tomas Hrycej, Siegfried Handschuh},     
    pdfkeywords={Residual Connection} {Deep Neural Network} {Convolutional Network} {Computer Vision}, 
    pdfnewwindow=true,      
    breaklinks=true,
}

\usepackage{SCITEPRESS}

\newcommand{\mnist}{\mbox{MNIST}}
\newcommand{\cifar}{\mbox{CIFAR-10}}

\DeclareMathOperator*{\softmax}{\text{Softmax}}

\begin{document}

\title{Reducing the Transformer Architecture to a Minimum}

\author{%
	\authorname{%
		Bernhard Bermeitinger\sup{\dagger}\sup{1},
		Tomas Hrycej\sup{\dagger}\sup{2},
		Massimo Pavone\sup{\ddagger}\sup{2},
		Julianus Kath\sup{\ddagger}\sup{2},
		and Siegfried Handschuh\sup{\dagger}\sup{2}
	}
	\affiliation{%
		\sup{1}Institute of Computer Science in Vorarlberg, University of St.Gallen (HSG), Dornbirn, Austria
	}
	\affiliation{%
		\sup{2}Institute of Computer Science, University of St.Gallen (HSG), St.Gallen, Switzerland
	}
	\email{%
        \sup{\dagger}firstname.lastname@unisg.ch,
        \sup{\ddagger}firstname.lastname@student.unisg.ch
    }
}


\abstract{%
	Transformers are a widespread and successful model architecture, particularly in Natural Language Processing~(NLP) and Computer Vision~(CV).
	The essential innovation of this architecture is the Attention Mechanism, which solves the problem of extracting relevant context information from long sequences in NLP and realistic scenes in CV\@.
	A classical neural network component, a Multi-Layer Perceptron~(MLP), complements the attention mechanism.
	Its necessity is frequently justified by its capability of modeling nonlinear relationships.
	However, the attention mechanism itself is nonlinear through its internal use of similarity measures.
	A possible hypothesis is that this nonlinearity is sufficient for modeling typical application problems.
	As the MLPs usually contain the most trainable parameters of the whole model, their omission would substantially reduce the parameter set size.
	Further components can also be reorganized to reduce the number of parameters.
	Under some conditions, query and key matrices can be collapsed into a single matrix of the same size.
	The same is true about value and projection matrices, which can also be omitted without eliminating the substance of the attention mechanism.
	Initially, the similarity measure was defined asymmetrically, with peculiar properties such as that a token is possibly dissimilar to itself.
	A possible symmetric definition requires only half of the parameters.
	All these parameter savings make sense only if the representational performance of the architecture is not significantly reduced.
	A comprehensive empirical proof for all important domains would be a huge task.
	We have laid the groundwork by testing widespread CV benchmarks: \mnist{}, \cifar{}, and, with restrictions, ImageNet.
	The tests have shown that simplified transformer architectures (a) without MLP, (b) with collapsed matrices, and (c) symmetric similarity matrices exhibit similar performance as the original architecture, saving up to \SI{90}{\percent} of parameters without hurting the classification performance.
}

\onecolumn \maketitle \normalsize \setcounter{footnote}{0} \vfill

\input{body.tex}

\bibliographystyle{apalike}
{\small \bibliography{references}}

\end{document}

%% file: body.tex
\section{\uppercase{Introduction}}\label{sec:introduction}
Recently, \emph{Large Language Models}~(LLMs) have shown impressive performance in producing complex text answers to given questions.
Their outstanding feature is the massive size of parameter sets (up to billions).
The rapidly growing parameter number has limited the possibility of developing such models (as well as objectively investigating their properties) to companies and institutions capable of making considerable investments in computing the model's parameters.

This is why it is of great interest to attempt to find more efficient configurations with fewer parameters without performance loss.
A computing model with an excellent success record is based on the transformer architecture~\cite{vaswani_attention_2017}.
Their success is due to an excellent ability to capture contextual information.
Initially developed for language processing, transformers have also been successfully used in Computer Vision (CV).
The analogy to language processing is the following: the semantics of individual words are determined by other words in the word sequence.
Frequently, the basic units are not words but tokens (e.g., $n$-grams consisting of $n$ consecutive letters).
Since the \emph{Vision Transformer}~\cite{dosovitskiy_image_2021}, in an image, the tokens are represented by \emph{patches} --- typically square regions of pixels in the image.
Other patches can influence or disambiguate a patch's conceptual meaning.
For example, the environment in which an individual object is embedded in the image may disambiguate the identification of a specific bird or mushroom species.

The fundamental concept of the transformer is that of \emph{attention}~\cite{bahdanau_neural_2016}.
It is based on the insight that a particular token's semantics are influenced by its close relationships with other tokens.
The tokens are encoded as real-valued vectors in a high-dimensional space (frequently around \num{1000} dimensions or more).
These vectors are called \emph{embeddings}.
The algebraic similarity between the embedding vectors measures the semantic proximity between the tokens.
This similarity measure is the vector product or the cosine angle between the vectors.
The weighting of tokens by such similarity measure is called attention, which, in analogy to human attention, focuses on relevant concepts.
From the computational point of view, a transformer is a structure consisting of
\begin{itemize}
	\item
	      an algorithm for consideration of token context, the \emph{attention mechanism}, and
	\item
	      a \emph{Multi-Layer Perceptron}~(MLP) for nonlinear transformation of intermediary data.
\end{itemize}

\paragraph{Multi-Head Attention}
For every transformer in the stack, the following processing is done by the attention mechanism (\emph{multi-head attention} or \emph{MHA}).
The input of a training sample in the stack's $s$-th Transformer (out of their total number $S$) is a sequence of input vectors $x_{si}$.
This sequence is transformed into an equally long sequence of output embeddings $z_{si}$.
Each of them is, for given weights, a formally linear transformation
\begin{equation}\label{eq:att_z}
	\begin{split}
		z_{si} & = \left( \sum^{i}_{j=1} a_{sij} x_{sj} W_{s}^V \right) W_s^O         \\
		       & = \left( \sum^{i}_{j=1} a_{sij} x_{sj}         \right) W_{s}^V W_s^O
	\end{split}
\end{equation}
i.e., a weighted average of input embeddings $x_{si}$, linearly transformed by matrix $ W_{s}^V W_s^O$.
The weight vectors $a_{si} = \left[ a_{si1}, a_{si2}, \dots, a_{sii} \right]$ are computed as
\begin{equation}\label{eq:att_a}
	a_{si} = \softmax \left( s_{si} \right)
\end{equation}
The vector argument of the $\softmax()$ function measures the similarity between a present token $x_Q$, \enquote{the query} and another token $x_K$, \enquote{the key}.
\begin{equation}\label{eq:att_s}
	s_{sij} = x_{si} W_{s}^Q W_{s}^{KT} x_{sj}^T
\end{equation}

This form of attention mechanism is referred to as \emph{single-head}.
A popular variant consists of an extension to multiple heads indexed by $h$:
\begin{equation} \label{eq:att_zh}
	z_{si} = \sum_{h=1}^H \left( \sum_{j=1}^i a_{shij} x_{sj} \right) W_{sh}^V W_{sh}^O
\end{equation}

Each head has its separate matrices $W_h^Q$, $W_h^K$, $W_h^V$, and $W_h^O$.
The weights are also computed separately as
\begin{equation}\label{eq:att_ah}
	a_{shi} = \softmax \left( s_{shi} \right)
\end{equation}
and
\begin{equation}\label{eq:att_sh}
	s_{shij} = x_{si} W_{sh}^Q W_{sh}^{KT} x_{sj}^T
\end{equation}

\paragraph{Multi-Layer Perceptron}
The second component is a standard MLP with a single hidden layer, applied to each intermediary embedding $z_{si}$:
\begin{equation} \label{eq:att_mlp}
	\begin{split}
		h_{si} & = f \left( z_{si} W_s^{(1)} + b_s^{(1)} \right) \\
		y_{si} & = h_{si} W_s^{(2)} + b_s^{(2)}
	\end{split}
\end{equation}
with $f()$ being a nonlinear function, usually the \emph{Gaussian Error Linear Unit (GELU)}~\cite{hendrycks_gaussian_2023}, weight matrices $W_s^{(1)}$ and $W_s^{(2)}$ as well as bias vectors $b_s^{(1)}$ and $b_s^{(2)}$.

\cite{he_simplifying_2024} have investigated the possibilities of simplifying the transformer architecture.
Their focus has been increasing the signal throughput through the network.
The proposed changes primarily consist of modifying or omitting shortcut connections and normalizing layers.
In addition, they have addressed the possibility of omitting matrices $W^V$ and $W^O$.
The last idea has also been implemented in our modifications proposed in~\cref{sec:single_head}.

Our focus is different: we intend to substantially reduce trainable parameters to accelerate the training and improve convergence.

\section{\uppercase{Transformer without the MLP}}\label{sec:no_mlp}
The MLP requires the majority of the parameters to be fitted.
This is justified by the argument that the MLP is the vehicle for implementing nonlinear mappings.

However, it can be argued that the first component, the attention mechanism, can also capture nonlinearities.
It is the variable weights that make the mapping nonlinear.
The argument of the $\softmax()$ function is already a quadratic function of input tokens, and the function itself is nonlinear.
Even if the $\softmax()$ were linear, the multiplication of input tokens by the weights $a_{sij}$ (which are quadratic in these tokens) would result in a cubic function of input tokens.
The nonlinearity of $\softmax()$ makes this mapping only more nonlinear.

So, a stack of $S$ transformers is a chain of $S$ at least cubic functions of the input, resulting in a function of polynomial order of at least $3S$.
This makes clear that subsequent processing by an MLP is not the only nonlinear element of the processing.
The extent of the task's nonlinearity cannot be assessed in advance.
Still, the hypothesis that a reduced transformer without an MLP may cover the nonlinearity needs for some tasks is justified and can be validated by appropriate tests.

Without the MLPs, the transformer architecture can be described in more explicit terms.
This is particularly the case if a single-head option is pursued.

\section{\uppercase{Single-head configuration}}\label{sec:single_head}
Although the matrices $W_s^Q$, $W_s^K$, $W_s^V$, and $W_s^O$ can theoretically map the embedding vector to an arbitrary vector width, it is common to keep this width constant throughout the model, referring to the \emph{model width} $N$.
Then, in the case of a single head, these matrices are square.
With square matrices, it is evident that $W_s^V W_s^O$ can be collapsed to a single matrix $W_s^{VO}$, and, analogically, $W_s^Q W_s^{KT}$ to $W_s^{QK}$.
This saves \SI{50}{\percent} of the attention module's parameters, from $4 S N^2$ to $2 S N^2$.

Concatenating the transformer-encoder layers without MLP leads to the following recursion:
\begin{equation} \label{eq:recursion_aij}
	\begin{split}
		y_{1i} & = \left(\sum_{j=1}^i a_{1ij} x_{1j} \right) W_1^{VO}                                              \\
		y_{2i} & = \left(\sum_{j=1}^i a_{2ij} y_{1j} \right) W_2^{VO}                                              \\
		       & = \left(\sum_{k=1}^i a_{2ik} \left( \sum_{j=1}^k a_{1kj} x_{1j} \right) W_1^{VO} \right) W_2^{VO} \\
		       & = W_1^{VO} W_2^{VO} \sum_{k=1}^i a_{2ik} \sum_{j=1}^k a_{1kj} x_{1j}                              \\
		       & \cdots
	\end{split}
\end{equation}

When stacking the attention modules, the matrices $W_s^{VO}$ concatenate to their product over $ s = 1, \ldots, S $.
Then, they collapse into a single matrix
\begin{equation} \label{eq:concatenated wvo}
	W^{VO} = \prod_{s=1}^{S} W_s^{VO}
\end{equation}

Since every sum $\sum_{j=1}^i a_{sij}$ is equal to unity (as a result of the softmax operation), every successive transformer layer performs a weighted mean of stacked inputs $x_{1j}$.

The total number of parameters with $ S $ matrices $W_s^{QK}$ and a single matrix $W^{VO}$ is $(S+1)N^2$, only slightly more than \SI{25}{\percent} of the original size without MLP\@.
So far, all this is possible without losing any expressive power of the single-head transformer without MLP --- only obsolete parameters are deleted.

In many NLP applications, the output of the last transformer of the stack is expected to produce an embedding of a word or a language token.
These output embeddings can be expected to come from the space spanned by the input words or tokens.
From this viewpoint, it may appear questionable to transform the input embeddings by matrices $W_s^{VO}$ and to re-transform them back into the word embeddings.
Then, it may be worth attempting to delete the value transformations.
This has also been the proposal of~\cite{he_simplifying_2024}, resulting in a simple weighted mean
\begin{equation}\label{eq:total_zij}
	z_{si} = \sum_{j=1}^i a_{sij} x_{sj}
\end{equation}

The output embedding $z_{Si}$ is a convex combination of input embeddings $x_{1i}$.
In other words, it is a member of the convex set spanned by $x_{1i}$.

This concept has been implemented in the Keras framework by setting the matrices $W_s^V$ and $W_s^O$ to unit matrices.
Collapsing $W_s^Q W_s^{KT}$ to $W_s^{QK}$ has been reached by setting the matrix $W^K$ to a unit matrix.
The newly defined matrix $W_s^{QK}$ replaces matrix $W_s^Q$.

\section{\uppercase{Multi-head configuration}}\label{sec:multi_head}
The relationships of~\cref{sec:single_head} are valid wherever the matrices $W_{sh}^V$, $W_{sh}^O$, $W_{sh}^Q$, and $W_{sh}^K$ are square.
This may also apply to multiple heads.
However, it is usual to commit to a reduced dimension per head.
With $H > 1$ heads, it is common to map the embedding vector to a narrower vector of width $\nicefrac{N}{H}$, assumed to be integer.

In such cases, the matrices $W_{sh}^V$, $W_{sh}^O$, $W_{sh}^Q$, and $W_{sh}^K$ are not square but of dimension $ \left( N, \nicefrac{N}{H} \right) $.
Collapsing $W_{sh}^Q W_{sh}^{KT}$ to $W_{sh}^{QK}$ is then no longer efficient since $W_{sh}^{QK}$ is of dimension $ (N, N) $ and has thus $N^2$ parameters while $W_{sh}^Q$ and $W_{sh}^K$ together have $\nicefrac{2N^2}{H}$, which is a smaller or equal number for $H>1$.

Moreover, it is impossible to equivalently concatenate the value/projection matrices $W_{sh}^{VO}$ to a unique product because of varying index $h$ along various paths through the heads.

Nevertheless, omitting the $W_{sh}^{VO}$ at all would have the same justification as for single-head configuration: the output embedding $z_{Si}$ would become a convex combination of input embeddings $x_{1i}$, which can be expected to correspond to a meaningful word or token.

\section{\uppercase{Symmetry of similarity}}\label{sec:symmetry}
The expression~\cref{eq:att_s} measures the similarity between queries and keys.
The general concept of characterizing similarity between vectors by their product is symmetric: $a$ is equally similar to $b$ as is $b$ to $a$.

However, the similarity between a key and a query evaluated with the help of $x_{si} W_{sh}^Q W_{sh}^{KT} x_{sj}^T$ is asymmetric.
This is because the matrices $W_{sh}^Q$ and $W_{sh}^K$ are potentially different.

This asymmetry leads to different similarities between $x_{si}$ and $x_{sj}$ in the roles of key and query: $x_{si}$ is not as similar to $x_{sj}$ as is $x_{sj}$ to $x_{si}$.
The vector $x_{si}$ is also not the most similar to itself.
The matrix product $W_{sh}^Q W_{sh}^{KT}$ is generally not positive definite, so it is not even guaranteed that the similarity of $x_{si}$ to itself is positive.

The asymmetry can be deliberate and justified from some viewpoints.
It is not a matter of course that the roles of queries and keys are symmetric.
However, some of the mentioned properties can make its use harmful.

The symmetry can be guaranteed by simply setting $W_{sh}^Q = W_{sh}^K$.
Then, half of the parameters dedicated to the query and key matrices can be economized.
In the single-head case, the same effect is reached by a symmetric matrix $W_s^{QK}$, with identical parameters mirrored over the diagonal, i.e., $w_{sij}^{QK} = w_{sji}^{QK}$.
Another possibility is to parameterize a lower triangular matrix $T_s^{QK}$ and to multiply it by its transpose, getting
\begin{equation}\label{eq:lower_diag}
	W_s^{QK} = T_s^{QK} T_s^{QKT}
\end{equation}
This amounts to the well-known \emph{Cholesky decomposition}~\cite{cholesky_note_1924} of a symmetric matrix.

With both methods, the number of parameters is $\frac{N(N+1)}{2}$ instead of $N^2$, or even $2N^2$ of the original version without collapsing $W^Q$ and $W^K$.

The symmetry is implemented by reusing $W_{sh}^Q$ as $W_{sh}^K$, omitting the use of $W_{sh}^K$ at all.

\section{\uppercase{Setup of computing experiments}}\label{sec:computing_experiments}
The benchmarks for the evaluation have been chosen from the CV domain.
They are medium-sized problems that can be run for a sufficient number of experiments.
This would not be possible with large models such as those used in language processing.

For the experiments, two well-known image classification datasets \mnist{}~\cite{lecun_gradient-based_1998} and \cifar{}~\cite{krizhevsky_learning_2009} were used.
\mnist{} contains grayscale images of handwritten digits (0--9) while \cifar{} contains color images of exclusively ten different mundane objects like \enquote{horse}, \enquote{ship}, or \enquote{dog}.
They contain \num{60 000} (\mnist{}) and \num{50 000} (\cifar{}) training examples.
Their respective preconfigured test split of each \num{10 000} examples are used as validation sets.
While \cifar{} is evenly distributed among all classes, \mnist{} can be considered almost equally distributed.

An important criterion is that the training set size is sufficient for good generalization.
The training size (as related to the number of model parameters) must be large enough for the model not to be underdetermined so that we can fairly assess the models' performances.
As a criterion for this, the overdetermination ratio of each benchmark candidate has been evaluated~\cite{hrycej_mathematical_2023}:
\begin{equation}\label{eq:q_coeff}
	Q = \frac{KM}{P}
\end{equation}
with $K$ being the number of training examples, $M$ being the output vector length (usually equal to the number of classes), and $P$ being the number of trainable model parameters.

This formula justifies itself by ensuring that the numerator $KM$ equals the number of constraints to be satisfied (the reference values for all training examples).
This number must be larger than the number of trainable parameters for the system to be sufficiently determined.
(Otherwise, there is an infinite number of solutions, most of which do not generalize.)
This is equivalent to the requirement for the overdetermination ratio $Q$ to be larger than unity.

\begin{table*}[t]
	\small
	\caption{%
		Results of 16 experiments on the two datasets \mnist{} and \cifar{} with 6 or 12 consecutive transformer encoders and 1 or 4 attention heads per encoder layer either with the default MLP inside each encoder layer or skipping it entirely.
		The loss and accuracy for the training and validation sets are reported after each model is trained for exactly \num{500} epochs.
	}
	\label{tab:results_with_and_without_mlp}
	\centering
	\begin{tabular}{l cc rrrrr}
		\toprule
		Dataset                 & \#Encs-\#Heads & MLP? & $Q$        & Train loss   & Val.\ loss   & Train.\ acc.\ [$\%$] & Val.\ acc.\ [$\%$] \\
		\midrule
		\multirow{8}{*}{\mnist} & 6-1            & yes  & \num{2.15} & \num{0.0067} & \num{0.0747} & \num{99.78}          & \num{98.38}        \\
		                        & 6-1            & no   & \num{6.46} & \num{0.0277} & \num{0.1023} & \num{99.07}          & \num{97.49}        \\
		                        & 6-4            & yes  & \num{2.09} & \num{0.0018} & \num{0.0739} & \num{99.95}          & \num{98.26}        \\
		                        & 6-4            & no   & \num{5.91} & \num{0.0021} & \num{0.0912} & \num{99.92}          & \num{98.29}        \\
		                        & 12-1           & yes  & \num{1.08} & \num{0.0052} & \num{0.0652} & \num{99.81}          & \num{98.71}        \\
		                        & 12-1           & no   & \num{3.29} & \num{0.0117} & \num{0.0970} & \num{99.62}          & \num{97.94}        \\
		                        & 12-4           & yes  & \num{1.08} & \num{0.0025} & \num{0.0656} & \num{99.92}          & \num{98.70}        \\
		                        & 12-4           & no   & \num{3.29} & \num{0.0026} & \num{0.1002} & \num{99.93}          & \num{98.10}        \\
		\midrule
		\multirow{8}{*}{\cifar} & 6-1            & yes  & \num{1.74} & \num{0.1533} & \num{2.2418} & \num{94.63}          & \num{60.24}        \\
		                        & 6-1            & no   & \num{4.93} & \num{0.9341} & \num{1.3590} & \num{66.16}          & \num{55.30}        \\
		                        & 6-4            & yes  & \num{1.74} & \num{0.1109} & \num{2.4033} & \num{96.01}          & \num{60.46}        \\
		                        & 6-4            & no   & \num{4.92} & \num{0.5621} & \num{1.6984} & \num{80.82}          & \num{52.37}        \\
		                        & 12-1           & yes  & \num{0.89} & \num{2.3026} & \num{2.3026} & \num{ 9.82}          & \num{10.00}        \\
		                        & 12-1           & no   & \num{2.52} & \num{0.5604} & \num{1.7219} & \num{79.48}          & \num{54.06}        \\
		                        & 12-4           & yes  & \num{0.89} & \num{0.0632} & \num{2.6379} & \num{97.92}          & \num{58.02}        \\
		                        & 12-4           & no   & \num{2.52} & \num{0.1787} & \num{2.3200} & \num{93.59}          & \num{55.60}        \\
		\bottomrule
	\end{tabular}
\end{table*}

The losses and accuracies in~\cref{tab:results_with_and_without_mlp} show that the performance with 12 encoders is not superior to that with 6 encoders.
The parameter set sizes with 12 encoders have been \num{563 242} with MLP and \num{198 100} without MLP.
This is substantially more than \num{287 686} and \num{101 470}, respectively, with 6 encoders.
Consequently, the latter variant has been adopted as a baseline.

\subsection{\uppercase{Results for \mnist}}\label{sec:computing_experiments_MNIST}
Following the arguments of~\cref{sec:no_mlp,sec:single_head,sec:multi_head,sec:symmetry}, the following reduced transformer variants have been tested:
\begin{itemize}
	\item
	      with and without an MLP in each transformer-encoder,
	\item
	      with 1 and 4 heads,
	\item
	      with the original matrix configuration as well matrix pair $W^Q$ and $W^K$ collapsed into one matrix, $W^V$ and $W^O$ omitted (one head variants only), and
	\item
	      with asymmetric and symmetric similarity measures.
\end{itemize}

\begin{table*}[t]
	\small
	\caption{%
		Loss and accuracy for different variants of transformer-encoder modifications on \mnist{}: 1 or 4 heads, with or without the MLP, with a single $W_{qk}$ matrix, no value and projection matrices, or a symmetric similarity measurement.
	}
	\label{tab:results_mnist_different_modifications}
	\centering
	\begin{tabular}{rcl rr rrrrr}
		\toprule
		\# Heads & MLP? & Modification & \# Parameters & $Q$         & Train loss   & Val.\ loss   & Train.\ acc.\ [$\%$] & Val.\ acc.\ [$\%$] \\
		\midrule
		1        & yes  & unchanged    & \num{279106}  & \num{2.15}  & \num{0.0067} & \num{0.0747} & \num{99.78}          & \num{98.38}        \\
		4        & yes  & unchanged    & \num{287746}  & \num{2.09}  & \num{0.0018} & \num{0.0739} & \num{99.95}          & \num{98.26}        \\
		1        & yes  & Wqk          & \num{257506}  & \num{2.33}  & \num{0.0037} & \num{0.0794} & \num{99.89}          & \num{98.43}        \\
		1        & yes  & Wqk+noWv,Vo  & \num{212866}  & \num{2.82}  & \num{0.0063} & \num{0.0951} & \num{99.78}          & \num{98.27}        \\
		1        & no   & unchanged    & \num{92890}   & \num{6.46}  & \num{0.0277} & \num{0.1023} & \num{99.07}          & \num{97.49}        \\
		4        & no   & unchanged    & \num{101530}  & \num{5.91}  & \num{0.0021} & \num{0.0912} & \num{99.92}          & \num{98.29}        \\
		1        & no   & symmetry     & \num{69910}   & \num{8.58}  & \num{0.0331} & \num{0.0783} & \num{98.85}          & \num{97.80}        \\
		4        & no   & symmetry     & \num{69910}   & \num{8.58}  & \num{0.0158} & \num{0.0762} & \num{99.46}          & \num{98.24}        \\
		1        & no   & Wqk          & \num{70570}   & \num{8.50}  & \num{0.0374} & \num{0.0996} & \num{98.70}          & \num{97.60}        \\
		1        & no   & Wqk+noWv,Vo  & \num{26650}   & \num{22.51} & \num{0.1697} & \num{0.1536} & \num{94.82}          & \num{95.32}        \\
		\bottomrule
	\end{tabular}
\end{table*}

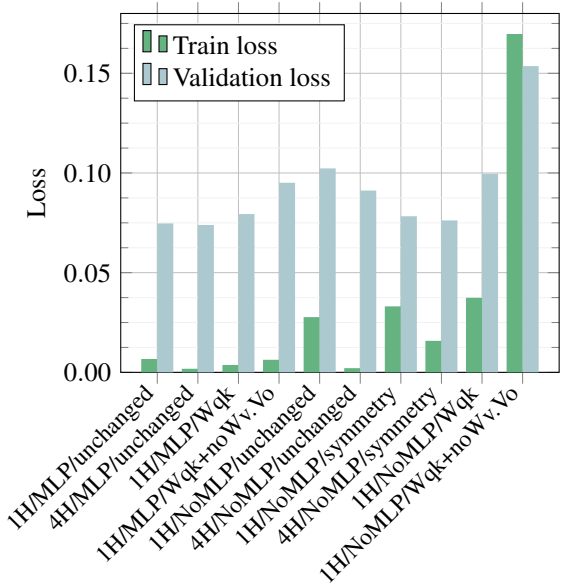
\begin{figure}
    \centering
    \tikzsetnextfilename{mnist_6e_row_loss}
    \begin{tikzpicture}[
        show background rectangle,
        tight background,
        background rectangle/.style={fill=white}
    ]
    \begin{axis}
    [
        width=.98\columnwidth,
        ybar=0pt,
        legend pos=north west,
        legend cell align=left,
        bar width=6pt,
        grid=both,
        grid style={line width=.1pt, draw=gray!10},
        major grid style={line width=.2pt,draw=gray!50},
        minor y tick num=3,
        ylabel={Loss},
        y label style={align=center},
        ymin=0, ymax=0.18,
        xtick=data,
        x tick label style={font=\footnotesize, anchor=east, rotate=45},
        symbolic x coords={
            {1H/MLP/unchanged},
            {4H/MLP/unchanged},
            {1H/MLP/Wqk},
            {1H/MLP/Wqk+noWv.Vo},
            {1H/NoMLP/unchanged},
            {4H/NoMLP/unchanged},
            {1H/NoMLP/symmetry},
            {4H/NoMLP/symmetry},
            {1H/NoMLP/Wqk},
            {1H/NoMLP/Wqk+noWv.Vo},
        },
        yticklabel style={%
            /pgf/number format/fixed,
            /pgf/number format/precision=2,
            /pgf/number format/fixed zerofill,
        },
    ]
        \addplot [draw=none, fill=HSG-Green!60] coordinates {
            ({1H/MLP/unchanged}, 0.0067)
            ({4H/MLP/unchanged}, 0.0018)
            ({1H/MLP/Wqk}, 00.0037)
            ({1H/MLP/Wqk+noWv.Vo}, 0.0063)
            ({1H/NoMLP/unchanged}, 0.0277)
            ({4H/NoMLP/unchanged}, 0.0021)
            ({1H/NoMLP/symmetry}, 0.0331)
            ({4H/NoMLP/symmetry}, 0.0158)
            ({1H/NoMLP/Wqk}, 0.0374)
            ({1H/NoMLP/Wqk+noWv.Vo}, 0.1697)
        };
        
        \addplot [draw=none, fill=HSG-Blue!60] coordinates {
            ({1H/MLP/unchanged}, 0.0747)
            ({4H/MLP/unchanged}, 0.0739)
            ({1H/MLP/Wqk}, 0.0794)
            ({1H/MLP/Wqk+noWv.Vo}, 0.0951)
            ({1H/NoMLP/unchanged}, 0.1023)
            ({4H/NoMLP/unchanged}, 0.0912)
            ({1H/NoMLP/symmetry}, 0.0783)
            ({4H/NoMLP/symmetry}, 0.0762)
            ({1H/NoMLP/Wqk}, 0.0996)
            ({1H/NoMLP/Wqk+noWv.Vo}, 0.1536)
        };

        \legend{{Train loss}, {Validation loss}}
    \end{axis}
    \end{tikzpicture}
	\caption{Training and validation losses attained by various reduced transformer-encoders with six encoder layers on \mnist{}\@.}\label{fig:MNIST_6E_row_loss}
\end{figure}

The variants depicted refer to the matrix options:
\begin{itemize}
	\item
	      \textit{unchanged} corresponds to the original attention module matrix variety;
	\item
	      \textit{Wqk} variants use a single matrix for the product $W^Q W^{KT}$; these variants are only available for a single attention head, and their similarity measure is asymmetric as in the original version;
	\item
	      \textit{noWv.Vo} denotes omitting the value matrices $W^V$ as well as the projection matrices $W^O$; also, these variants imply a single attention head and asymmetric similarity measurement;
	\item
	      \textit{symmetric} variants are committed to symmetric similarity measures; $W^V$ and $W^O$ are left untouched.
\end{itemize}

The performances of the individual variants are given in~\cref{tab:results_mnist_different_modifications}.
For better comparability, the losses are additionally depicted in~\cref{fig:MNIST_6E_row_loss}.

The following observations can be made:
\begin{itemize}
	\item
	      The original variants with MLPs perform better than those without MLPs on the training set.
	\item
	      By contrast, their advance disappears on the validation set, particularly if the symmetric similarity metrics are used.
	\item
	      The variant with asymmetric similarity without MLP is inferior to the analogical one with symmetric similarity.
	\item
	      The minimum variant with query and key matrices $W^Q, W^{K}$ collapsed to $W^{QK} = W^Q W^{KT}$ and additionally omitted value and projection matrices show a higher loss than other variants.
	      This may be due to its dramatically reduced parameter number, which may lead to an insufficient capacity to capture nonlinearities.
\end{itemize}

As \mnist{} is a relatively easy benchmark, the accuracy results are very close to each other.
The parameter numbers are substantially different.
The symmetric variant without MLP has only about \SI{25}{\percent} of the parameter number of the original, full variant with MLP.
The variant with collapsed matrices has about \SI{33}{\percent} of the original parameters.
The parameters include, in addition to the attention modules of all transformer-encoders, the embedding matrix reducing the image patch to the embedding vector.

The number of parameters has a strong effect on the generalization capability of the model.
This can be quantified with the help of the overdetermination ratio from~\cref{eq:q_coeff} in column $Q$ of~\cref{tab:results_mnist_different_modifications}.
The loss gap between the training and validation sets is the largest for the original version with $Q$ close to unity while it shrinks towards the symmetric version without MLPs.

\subsection{\uppercase{Results for \cifar}}\label{sec:computing_experiments_Cifar10}
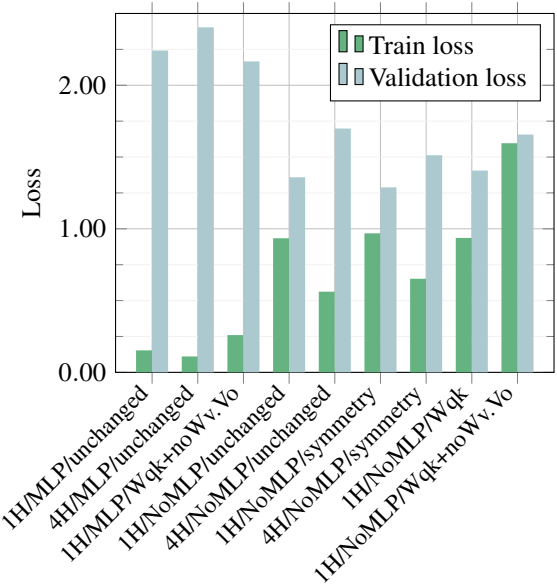
\begin{figure}
    \centering
    \tikzsetnextfilename{cifar_6e_row_loss}
    \begin{tikzpicture}[
        show background rectangle,
        tight background,
        background rectangle/.style={fill=white}
    ]
    \begin{axis}
    [
        width=0.98\columnwidth,
        ybar=0pt,
        legend pos=north east,
        legend cell align=left,
        bar width=6pt,
        grid=both,
        grid style={line width=.1pt, draw=gray!10},
        major grid style={line width=.2pt,draw=gray!50},
        minor y tick num=3,
        ylabel={Loss},
        y label style={align=center},
        ymin=0, ymax=2.5,
        xtick=data,
        x tick label style={font=\footnotesize, anchor=east, rotate=45},
        symbolic x coords={
            {1H/MLP/unchanged},
            {4H/MLP/unchanged},
            {1H/MLP/Wqk+noWv.Vo},
            {1H/NoMLP/unchanged},
            {4H/NoMLP/unchanged},
            {1H/NoMLP/symmetry},
            {4H/NoMLP/symmetry},
            {1H/NoMLP/Wqk},
            {1H/NoMLP/Wqk+noWv.Vo},
        },
        yticklabel style={%
            /pgf/number format/fixed,
            /pgf/number format/precision=2,
            /pgf/number format/fixed zerofill,
        },
    ]
        \addplot [draw=none, fill=HSG-Green!60] coordinates {
            ({1H/MLP/unchanged}, 0.1533)
            ({4H/MLP/unchanged}, 0.1109)
            ({1H/MLP/Wqk+noWv.Vo}, 0.2597)
            ({1H/NoMLP/unchanged}, 0.9341)
            ({4H/NoMLP/unchanged}, 0.5621)
            ({1H/NoMLP/symmetry}, 0.9686)
            ({4H/NoMLP/symmetry}, 0.6521)
            ({1H/NoMLP/Wqk}, 0.9364)
            ({1H/NoMLP/Wqk+noWv.Vo}, 1.5961)
        };
        
        \addplot [draw=none, fill=HSG-Blue!60] coordinates {
            ({1H/MLP/unchanged}, 2.2418)
            ({4H/MLP/unchanged}, 2.4033)
            ({1H/MLP/Wqk+noWv.Vo}, 2.1659)
            ({1H/NoMLP/unchanged}, 1.3590)
            ({4H/NoMLP/unchanged}, 1.6984)
            ({1H/NoMLP/symmetry}, 1.2885)
            ({4H/NoMLP/symmetry}, 1.5125)
            ({1H/NoMLP/Wqk}, 1.4057)
            ({1H/NoMLP/Wqk+noWv.Vo}, 1.6565)
        };

        \legend{{Train loss}, {Validation loss}}
    \end{axis}
    \end{tikzpicture}
	\caption{Training and validation losses attained by various reduced transformer-encoders with six encoder layers on \cifar{}\@.}\label{fig:Cifar10_6E_row_loss}
\end{figure}

The variants tested are analogical to those for \mnist{}\@.
The losses and accuracies attained after \num{500} epochs are given in~\cref{tab:results_cifar10_different_modifications}, the losses additionally
in~\cref{fig:Cifar10_6E_row_loss}.

\begin{table*}[t]
	\small
	\caption{%
		Loss and accuracy for different variants of transformer-encoder modifications on \cifar{}: 1 or 4 heads, with or without MLP, with a single $W_{qk}$ matrix, no value and projection matrices, or a symmetric similarity measurement.
	}
	\centering
	\begin{tabular}{rcl rr rrrrr}
		\toprule
		\# Heads & MLP? & Modification & \# Parameters & $Q$         & Train loss   & Val.\ loss   & Train.\ acc.\ [$\%$] & Val.\ acc.\ [$\%$] \\
		\midrule
		\num{1}  & yes  & unchanged    & \num{287686}  & \num{1.74}  & \num{0.1533} & \num{2.2418} & \num{94.63}          & \num{60.24}        \\
		\num{4}  & yes  & unchanged    & \num{287746}  & \num{1.74}  & \num{0.1109} & \num{2.4033} & \num{96.01}          & \num{60.46}        \\
		\num{1}  & yes  & Wqk+noWv,Vo  & \num{221446}  & \num{2.26}  & \num{0.2597} & \num{2.1659} & \num{90.53}          & \num{54.98}        \\
		\num{1}  & no   & unchanged    & \num{101470}  & \num{4.93}  & \num{0.9341} & \num{1.3590} & \num{66.16}          & \num{55.30}        \\
		\num{4}  & no   & unchanged    & \num{101530}  & \num{4.92}  & \num{0.5621} & \num{1.6984} & \num{80.82}          & \num{52.37}        \\
		\num{1}  & no   & symmetry     & \num{78490}   & \num{6.37}  & \num{0.9686} & \num{1.2885} & \num{64.80}          & \num{55.85}        \\
		\num{4}  & no   & symmetry     & \num{78490}   & \num{6.37}  & \num{0.6521} & \num{1.5125} & \num{76.10}          & \num{55.52}        \\
		\num{1}  & no   & Wqk          & \num{79150}   & \num{6.32}  & \num{0.9364} & \num{1.4057} & \num{66.03}          & \num{53.70}        \\
		\num{1}  & no   & Wqk+noWv,Vo  & \num{35230}   & \num{14.19} & \num{1.5961} & \num{1.6565} & \num{40.52}          & \num{39.17}        \\
		\bottomrule
	\end{tabular}
	\label{tab:results_cifar10_different_modifications}
\end{table*}

The result characteristics are similar to those for \mnist{} but more distinct:
\begin{itemize}
	\item
	      The original variant with MLP reaches the best training set loss but the worst validation set loss.
	\item
	      Compared to the original variant, the reduced variants without MLP and with symmetric similarity are superior in generalization.
	\item
	      This also applies to the variant with collapsed key and query matrices.
	\item
	      Even the minimum variant with all considered matrix reductions (except for symmetry), whose parameter count is only a tenth of the original version with MLP, shows a better validation set performance than the original variant with all matrices and MLP\@.
\end{itemize}

The measured accuracies are roughly consistent with the losses on the training set.
On the validation set, some of them follow, paradoxically, a different ranking.
However, the fact that the loss, not the accuracy, is explicitly trained justifies the arguments via loss rather than accuracy.

\subsection{\uppercase{Trials with ImageNet}}\label{sec:computing_experiments_ImageNet}
Several trials on the ImageNet dataset~\cite{russakovsky_imagenet_2015} have been conducted to support the hypotheses with a larger benchmark.
Unfortunately, the baseline run with the original transformer architecture, including MLP, has not been successful.
In all trials, \emph{Adam} failed to find a substantial improvement in the initial parameter state.
By contrast, without MLP, it has been converging at least to a state with a moderate classification performance.
This is why we cannot present a serious study on ImageNet.
It can only be concluded that discarding MLP is helpful for convergence.
The proof that this variant's performance is acceptable is still pending, and further work will be required to provide it.

\section{\uppercase{Conclusions and Limitations}}\label{sec:conclusions}
The experiments presented have shown limited utility of some parameter-extensive components of the transformer architecture.
In particular, the following findings can be formulated:
\begin{itemize}
	\item
	      The MLP component is frequently presented as necessary for capturing nonlinearities in the modeled relationship.
	      However, the inherent nonlinearity of the similarity measures seems powerful enough in many practical cases.
	\item
	      While the classification performance without the MLPs is not significantly inferior to that with MLPs, a substantial benefit is saving the parameters.
	      With model size $ N $, the attention mechanism requires $ 4 N^2 $ parameters in the form of matrices $ W^Q $, $ W^K W^V $, and $W^O$.
	      The size of the MLP is usually chosen as an integer multiple of $h$ of the model size.
	      Then, the MLP consists of weights and biases of two layers, with a total of $h N (N + 1) + N (h N + 1) = 2 h N^2 + hN + N \approx 2 h N^2 $.
	      If the multiple is $ h = 4 $, MLP has double the number of parameters as the attention mechanism.
	      Consequently, omitting MLP reduces the parameters to \SI{33}{\percent} of the original size.
	\item
	      Symmetric similarity measures tend to perform better than asymmetric ones, with \SI{50}{\percent} fewer query and key matrix parameters.
	      This improvement may be reached by excluding undesirable freedoms, such as a token being dissimilar to itself.
	      The parameter reduction can be expected to constrain the search for the optimum fit fruitfully.
	\item
	      Collapsing the value and the key matrix into one is another possibility of reducing the parameter set of these matrices by \SI{50}{\percent}.
	\item
	      Omitting the value matrix $W^V$ and the projection matrix $W^O$ reduces the parameters of the whole attention module by \SI{50}{\percent}.
	      This variant has also been proposed by~\cite{he_simplifying_2024}, with the observation of no significant performance loss in NLP benchmarks.
	\item
	      Both preceding reductions amount to a reduction to \SI{25}{\percent} of the original attention module size.
	\item
	      In our experiments, the variants with the collapsed query/key matrices, omitted value, and projection matrices are slightly inferior for \mnist{} but equal for \cifar{}.
	      These minimum variants have less than \SI{10}{\percent} of parameters compared with the classical transformers, including MLP.
	      Compared to the architecture with 12 encoders, it is as little as \SI{5}{\percent}.
\end{itemize}
The savings in computing time have been proportional to the savings in parameter numbers.

Our research has been limited to image processing benchmarks \mnist{}, \cifar{}, and ImageNet.
The experiments with the last benchmark have partially failed due to computing problems.
Empirical evidence with the help of two medium-sized benchmarks and an incomplete test of a larger one is not satisfactory.
This requests further research with more robust algorithms.
There is considerable potential for second-order optimization methods such as the conjugate gradient algorithm of ~\cite{fletcher_function_1964}, thoroughly described in~\cite{press_numerical_1992}.
This algorithm's convergence is excellent, but implementing the stopping rule in widespread packages seems to improve its ability to prevent early stops before reaching the minimum region.

Limitations to image processing suggest further extension.
The proper domain of transformers is NLP\@.
An obstacle to its investigation is the size of benchmark problems, so most published investigations consist of observing the performance of fine-tuning pre-trained models.
To use pre-trained parameter sets, these fine-tuned models must be identical or almost identical to the pre-trained models.
This makes the testing of different architectures difficult.
A possibility is to use a large model used for pre-training as a \emph{teacher} and a medium-sized model as \emph{student}, mimicking its performance.
This procedure, referred to as \emph{knowledge distillation}, has been proposed by~\cite{hinton_distilling_2015} and used, e.g., by~\cite{sun_patient_2019}.

These will be important focuses soon.